\documentclass{nextgame2026}

\PassOptionsToPackage{numbers,compress}{natbib}

\usepackage{microtype}
\usepackage{graphicx}
\usepackage{booktabs}
\usepackage{hyperref}
\usepackage{amsmath}
\usepackage{amssymb}
\usepackage{caption}

\title[Translation Gap in Cooperative MARL]
{Learned Coordination Conventions in Cooperative MARL:
Measuring the Translation Gap Between Theory-Informed Roles and Learned Routing}

\optauthor{%
\Name{Yoosung Hong} \Email{yoosunghong.main@gmail.com}\\
\addr Independent Researcher
}

\begin{document}

\maketitle


\begin{abstract}
Role-semantic assignments supply priors over how heterogeneous agents may coordinate; cooperative MARL agents instead settle on conventions through decentralized, non-stationary learning, with no guarantee the resulting structure matches those priors -- an equilibrium selection question that theory alone cannot resolve. We ask whether the convention selected by training is legible from the policy architecture, and how it relates to a theory-informed role-semantic prior, using a diagnostic that combines a role-routing matrix, formation sensitivity ($\Delta_{\max}$), and gradient/occlusion attribution across three-role MiniGrid and SMACv2 (Terran), five MAPPO conditions, and a 3v3--9v9 scaling study.

Label-conditioned attention produces substantially more concentrated and role-specific routing than flat MLP baselines (entry std $0.246$ vs.\ $0.055$); the signature is stable under scaling, transfers zero-shot from 3v3 to 9v9 above the from-scratch baseline ($0.224$ vs.\ $0.110$), and is invariant to ally-slot padding. Removing role labels under shared attention costs 14.0\,pp. Under 5-seed re-evaluation, three of four theory-informed predictions hold and one is tied; the earlier 3-seed Strike$\to$Vanguard finding was a small-$n$ artifact, showing the diagnostic is itself seed-sensitive. We read this as evidence that architecture-exposed routing makes aspects of learned conventions measurable, contributing an empirical diagnostic for cooperative MARL rather than a new solution concept.
\end{abstract}

\section{Introduction}
\label{sec:introduction}

A central question for game-theoretic learning is no longer only \emph{what equilibrium should exist}, but \emph{what coordination convention a learning system actually settles on} -- an equilibrium selection question that theory alone cannot resolve. Classical role-based reasoning supplies strategic priors under explicit rationality assumptions; cooperative MARL agents instead select a convention through decentralized gradient updates in non-stationary environments, with no a priori reason the resulting role-conditioned structure should align with those priors -- what we term a translation gap between theory-informed expectations and learned coordination structure.

We treat this as a question of coordination legibility: whether the role-conditioned interaction structure of a trained policy is measurable from its architecture, and how it relates to a theory-informed role-semantic prior. When role labels are injected into both \texttt{self\_tok} and \texttt{ally\_tok}, cross-attention exposes role-pair patterns to direct inspection. Our central claim is that this pathway yields a more concentrated, role-specific signature than flat MLP baselines and is robust to scaling, cross-scale transfer, and padding -- a measurable architectural property of the learned convention, not a claim that attention causally explains behavior or that the convention is an equilibrium.

Contributions: (i) a diagnostic framework—role-routing matrix, $\Delta_{\max}$, and gradient/occlusion analysis—for measuring learned coordination conventions; (ii) evidence on MiniGrid and SMACv2 Terran, across 3v3–9v9, that label-conditioned attention yields more concentrated routing, stronger zero-shot transfer, and padding-invariant behavior than MLP baselines; (iii) a 5-seed re-evaluation showing partial prior alignment and exposing small-$n$ seed artifacts.
\section{Related Work}
\label{sec:related}

Role-based MARL learns or decomposes roles to structure cooperation: ROMA~\cite{roma2020} learns latent roles and RODE~\cite{rode2021} decomposes tasks by role, with related lines on committed exploration, shared experience, and selective parameter sharing~\cite{maven2019,entitycentric,maac_role}; we instead analyze how architectures exploit \emph{pre-defined} role labels. On the architecture side, self-attention~\cite{attention2017,zambaldi2019} provides a relational inductive bias for structured entity inputs, UPDeT~\cite{updet2021} uses transformers for transfer, MAAC~\cite{maattn} uses attention for multi-agent actor-critic learning, and MADDPG~\cite{maddpg} uses centralized training for multi-agent coordination; these methods do not provide a role-conditioned entity-selection path, while cooperative optimizers such as QMIX~\cite{qmix} and MAPPO~\cite{mappo} lack a structural path for role-pair routing. Work on convention and equilibrium selection -- Other-Play~\cite{otherplay2020}, Off-Belief Learning~\cite{obl2021}, and human-AI coordination~\cite{overcooked2019} -- shapes \emph{which} convention is learned; we instead hold the algorithm fixed and ask whether the selected convention becomes \emph{inspectable}, and how it relates to a role-semantic designer prior. Methodologically this places us in explainable RL~\cite{xrl_survey2024}, and we treat attention weights cautiously given the ongoing debate on whether they constitute explanation~\cite{jain2019attention,wiegreffe2019attention}.

\section{Mechanism: Role-Label-Conditioned Routing}
\label{sec:theory}

\paragraph{Routing path.} In an MLP, role labels are concatenated with other features and ``which entity to condition on given my role'' is implicit in weight products -- a weaker structural prior, not an impossibility. In our architecture, role labels appear in both \texttt{self\_tok[4:7]} (encoded into the Query) and each \texttt{ally\_tok} role-onehot block (encoded into Key/Value), so the cross-attention score $\text{softmax}((W_Q \mathbf{x}_i)^\top(W_K \mathbf{x}_j)/\sqrt{d_k})$ becomes a direct function of the role-pair $(r_i,r_j)$. Cross-attention is applied per entity group (ally, enemy, zone) with the self-embedding as query; our routing analysis isolates the ally branch. An Intra-Set Self-Attention block enriches each ally token before the cross-attention Key. We treat these weights as architecture-exposed coordination patterns; attribution and occlusion analyses (\S\ref{sec:experiments}) provide complementary, less interpretation-laden views.

\paragraph{Why these predictions?} Under the designer-specified role semantics, Strike (ranged damage) should benefit most from Support, Vanguard (frontline) should react most to formation changes, and Support should distribute attention more uniformly while tracking team state.

\paragraph{Testable predictions.} (1) attention produces more concentrated role routing than MLP -- routing-matrix entry std at least $2\times$ the MLP gradient-attribution std \textbf{[PASS, $4.5\times$: $0.246$ vs.\ $0.055$, 5 seeds]}; (2) under shared-attention, zeroing role labels degrades performance by $>$5\,pp \textbf{[PASS, $-14.0$\,pp; bundles label removal with policy sharing -- upper-bounds rather than isolates the label effect]}; (3) $\Delta_{\max}$ is highest for Vanguard \textbf{[TIED: Support $0.074$, Vanguard $0.066$, gap $<$ either seed std; qualitatively recovers ``non-DPS roles are formation-sensitive'']}; (4) Strike attends preferentially to Support \textbf{[PASS in $4/5$ seeds at both 3v3 and 9v9; the 3-seed Strike$\to$Vanguard result was a small-$n$ artifact]}.

\section{Method}
\label{sec:method}

\paragraph{Environments.} \textbf{Domain~1} is a custom $15{\times}15$ MiniGrid~\cite{minigrid2023} task with three role-specialized agents (Strike, Vanguard, Support) cooperating to capture and hold zones under role-specific reward shaping. \textbf{Domain~2} is a custom SMACv2~\cite{smac2019,smacv2} scenario using three Terran unit types as role proxies (Marine = Strike, Marauder = Vanguard, Medivac = Support). SMACv2 does not natively expose role labels; we inject a 3-dim unit-type one-hot into both the agent's own token and each ally token, preserving the structural condition required for role-conditioned Q/K routing.

\paragraph{Entity-centric observation.} Agents receive four named token groups -- \texttt{self\_tok}, \texttt{ally\_tok}, \texttt{enemy\_tok}, \texttt{zone\_tok} -- plus validity masks. The role one-hot (3 dims) appears in both \texttt{self\_tok} and each \texttt{ally\_tok} slot; this placement is load-bearing for the routing mechanism. The 3v3 SMACv2 observation is 65-dim (182-dim padded for the scaling study).

\begin{figure}[t]
  \centering
  \includegraphics[width=0.6\textwidth]{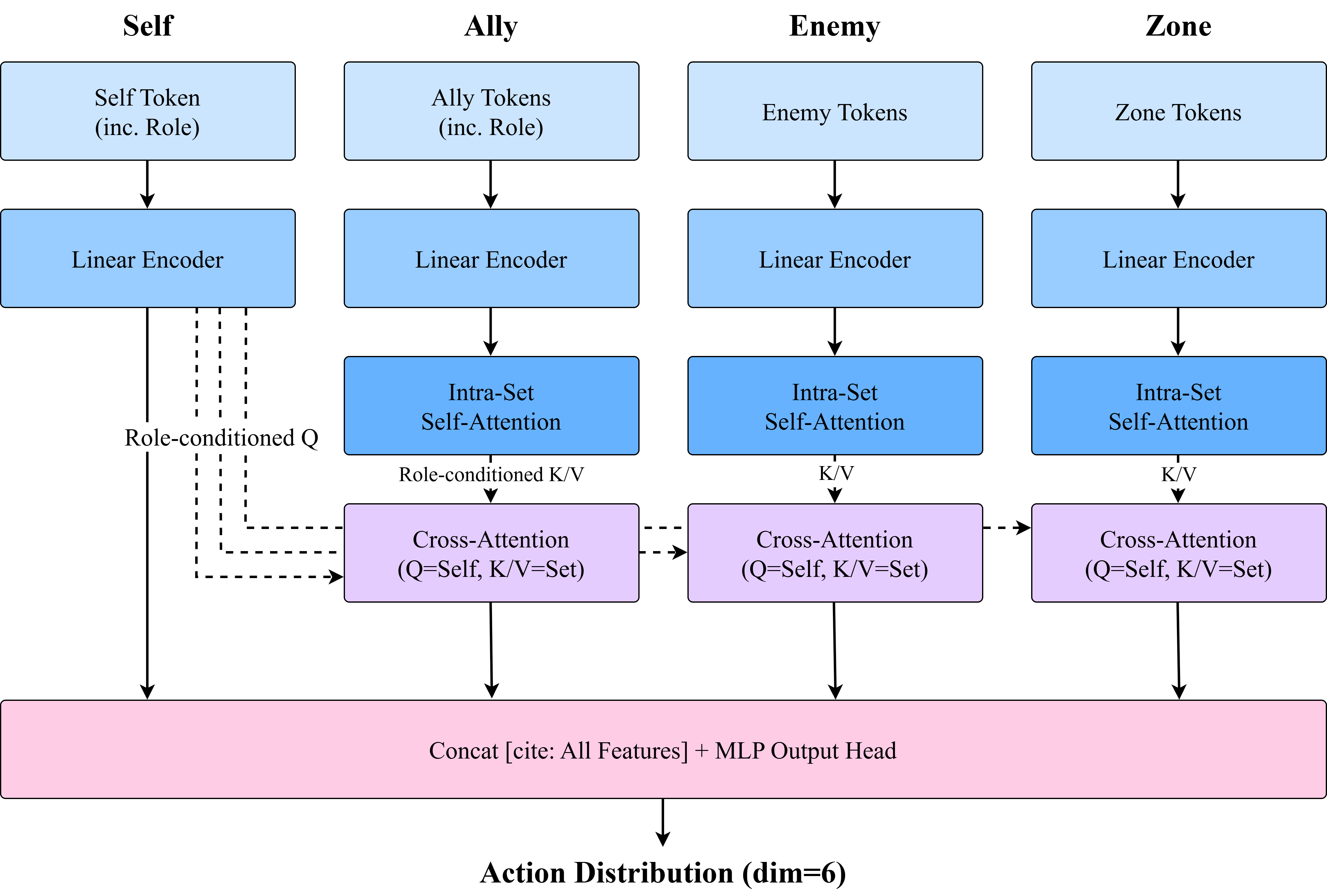}
  \caption{SLIC architecture (Cond.~1). Per-set encoders feed intra-set self-attention and self-query cross-attention; dashed arrows mark the role-conditioned routing path.}
  \label{fig:architecture}
\end{figure}

\paragraph{Five ablation conditions.} The conditions span two axes: \emph{architecture} (attention vs.\ MLP) and \emph{role-label + policy sharing}. \textbf{C1 Full}: per-set encoders (dim 64) + Intra-Set Self-Attn (residual+LN) + Cross-Attn with self as query; per-role policies, labels present. \textbf{C2}: C1 minus Intra-Set blocks. \textbf{C3 Role+MLP}: flat MLP, per-role, labels in input. \textbf{C4 No-Role (shared attention)}: C1 architecture, role one-hots zeroed, single shared policy. \textbf{C5 Flat MLP}: C3 with labels zeroed. C1 vs.\ C4 bundles label removal with a per-role$\to$shared-policy change (upper bound on label effect); C3 vs.\ C5 isolates label-in-input for MLP ($+1.6$\,pp, within noise); C4 vs.\ C5 confounds both axes plus a $3{\times}$ data-per-parameter advantage from sharing. All attention applies \texttt{key\_padding\_mask} with slot-0 force-unmask safety. Training uses MAPPO~\cite{mappo} (PPO~\cite{ppo2017} with a centralized critic) and identical hyperparameters (2M steps; 5 seeds for all 3v3 conditions and 9v9 C1/C3, 3 seeds for 6v6 C1/C3; GAE~\cite{gae2016} $\lambda{=}0.95$, $\gamma{=}0.99$, clip $0.2$, entropy $0.01$, Adam $3{\times}10^{-4}$, batch 512); reward mix $r=0.8 r_{\text{ind}}+0.2 \bar r_{\text{team}}$.

\section{Experiments}
\label{sec:experiments}

\subsection{Track 1: Performance Comparison (SMACv2 3v3)}

\begin{table}[t]
\centering\small
\setlength{\tabcolsep}{6pt}
\caption{Track 1 win rate on SMACv2 3v3 (mean$\pm$std over 5 seeds of the last-10 evaluation average). \textbf{Bold} marks the highest mean.}
\label{tab:track1}
\begin{tabular}{lccccc}
\toprule
 & C1: Full & C2: No Self-Attn & C3: Role+MLP & C4: No Role (shared) & C5: Flat MLP \\
\midrule
Win rate & $\mathbf{0.794{\pm}0.033}$ & $0.656{\pm}0.104$ & $0.480{\pm}0.155$ & $0.654{\pm}0.087$ & $0.464{\pm}0.049$ \\
$n$ seeds & 5 & 5 & 5 & 5 & 5 \\
\bottomrule
\end{tabular}
\end{table}

Three observations matter. Attention helps: C1 outperforms C3 by 31.4\,pp. Intra-set context helps: C2 trails C1 by 13.8\,pp. Removing labels under shared-attention is associated with a 14.0\,pp drop (C4 vs.\ C1; a label$+$sharing confound that upper-bounds the label effect, \S\ref{sec:method}). C3 and C5 are statistically indistinguishable ($0.480$ vs.\ $0.464$): adding labels to an MLP does not by itself produce useful role-conditioned routing. Expanding all 3v3 conditions to 5 seeds preserves the sign of every contrast.

\subsection{Scaling Study: 3v3, 6v6, 9v9}
\label{sec:scaling}

We test whether the C1 vs.\ C3 contrast survives larger team sizes. We train balanced 3v3, 6v6, and 9v9 variants with equal Strike/Vanguard/Support counts; enemy composition scales in parallel while preserving the asymmetric design (no enemy Support unit). All scaling runs use identical hyperparameters and a centralized critic over concatenated global observations.

\begin{table}[t]
\centering
\small
\setlength{\tabcolsep}{7pt}
\caption{Scaling results (mean$\pm$std). 3v3 and 9v9 now use 5 seeds for both C1 and C3; 6v6 remains 3 seeds.}
\resizebox{0.75\linewidth}{!}{%
\label{tab:scaling}
\begin{tabular}{lcccc}
\toprule
Scale & C1 Last-10 & C3 Last-10 & C1 AUC & C3 AUC \\
\midrule
3v3 & $\mathbf{0.794{\pm}0.033}$ & $0.480{\pm}0.155$ & $\mathbf{0.678{\pm}0.019}$ & $0.328{\pm}0.059$ \\
6v6 & $\mathbf{0.380{\pm}0.098}$ & $0.363{\pm}0.144$ & $\mathbf{0.379{\pm}0.064}$ & $0.311{\pm}0.098$ \\
9v9 & $\mathbf{0.110{\pm}0.090}$ & $0.060{\pm}0.032$ & $\mathbf{0.119{\pm}0.067}$ & $0.089{\pm}0.052$ \\
\bottomrule
\end{tabular}
}
\end{table}

In 3v3, C1 is clearly better than C3. In 6v6/9v9 the full model remains ahead on AUC and late-training averages, but margins shrink and seed variance grows -- the 6v6 ($0.380$ vs.\ $0.363$) and 9v9 ($0.110$ vs.\ $0.060$) gaps are within or close to seed std and we do not claim significance. The performance claim is therefore modest: attention improves small-team performance and shows a noisy directional advantage at scale. The structural claim is stronger: role-conditioned routing persists at 9v9 (read from off-diagonal contrasts as magnitudes compress with more slots), cross-scale transfer remains positive, and the model is invariant to padded slots.

\paragraph{Cross-scale zero-shot transfer.} C1's policy is invariant to ally count by construction. A 3v3-trained C1 checkpoint evaluated zero-shot on 9v9 (50 episodes per seed, 5 seeds) reaches $0.224\pm 0.088$ -- above from-scratch 9v9 C1 ($0.110\pm 0.090$) and far above 9v9 C3 ($0.060\pm 0.032$). C3 transfers poorly throughout ($\sim 0.084$), consistent with MLP weights memorizing a fixed slot pattern.

\begin{figure}[t]
\centering
\begin{minipage}[t]{0.47\linewidth}
  \vspace{0pt}
  \centering\small
  \setlength{\tabcolsep}{5pt}
  \captionof{table}{Zero-shot transfer of 3v3-trained checkpoints, evaluated without retraining (50 fresh episodes per seed; mean$\pm$std over 5 seeds). The 3v3 row differs slightly from Table~\ref{tab:track1}, which reports the training-time last-10 average rather than fresh rollouts.}
  \label{tab:transfer}
  \begin{tabular}{lccc}
  \toprule
  Target & Cond.~1 & Cond.~3 & Gap \\
  \midrule
  3v3 & $\mathbf{0.796{\pm}0.054}$ & $0.468{\pm}0.153$ & 32.8 pp \\
  6v6 & $\mathbf{0.212{\pm}0.046}$ & $0.084{\pm}0.033$ & 12.8 pp \\
  9v9 & $\mathbf{0.224{\pm}0.088}$ & $0.084{\pm}0.043$ & 14.0 pp \\
  \bottomrule
  \end{tabular}
\end{minipage}
\hfill
\begin{minipage}[t]{0.49\linewidth}
  \vspace{0pt}
  \centering\small
  \setlength{\tabcolsep}{5pt}
  \captionof{table}{Padding-confound ablation at 3v3 (5 seeds, 30 episodes per seed). KL is divergence from the zero-fill baseline.}
  \label{tab:padding}
  \begin{tabular}{lcc}
  \toprule
  Fill strategy & C3 KL & C3 win rate \\
  \midrule
  zeros & 0.000 & $0.460{\pm}0.138$ \\
  noise $\sigma{=}0.05$ & 0.007 & $0.500{\pm}0.078$ \\
  noise $\sigma{=}0.20$ & 0.136 & $\mathbf{0.567{\pm}0.127}$ \\
  mean valid ally & 0.096 & $0.407{\pm}0.171$ \\
  copy valid ally & 0.101 & $0.453{\pm}0.209$ \\
  \bottomrule
  \end{tabular}
\end{minipage}
\end{figure}

\paragraph{Padding-confound ablation.} A potential confound for the 3v3 gap is that C3 must process zero-padded ally slots that C1's mask blocks. Across fill strategies (Table~\ref{tab:padding}), additional C1 checks show that the masked-attention policy is bit-exact invariant for data-driven fills and changes only negligibly under noise (KL $\leq 0.021$). Table~\ref{tab:padding} therefore reports the C3 sensitivity, where responses are non-monotonic. The best C3 fill recovers $+10.7$,pp of the $31.4$,pp gap, bounding the padding contribution above by $\sim 34\%$ and leaving the majority as architectural residual.

\subsection{Track 2: Role-to-Role Routing}

\paragraph{Formation sensitivity and routing matrix.} Two fixed scenarios -- \textit{Clustered} (allies near one landmark) and \textit{Spread} (each ally at a distinct landmark) -- define $\Delta_{\max}$ per role as the maximum difference in ally-branch cross-attention weights. C1 with 5 seeds yields $\Delta_{\max}=0.024{\pm}0.017, 0.066{\pm}0.039, 0.074{\pm}0.029$ for Strike, Vanguard, Support: Support and Vanguard are statistically tied at the top, both well above Strike. Prediction~3 (Vanguard strictly highest) is therefore not strictly supported in mean ordering ($\Delta_\text{Support} > \Delta_\text{Vanguard}$ by $0.008$, smaller than either std), but the qualitative design intent -- the two non-DPS roles are formation-sensitive while Strike is not -- is recovered. The routing matrix (Table~\ref{tab:routing}) extracts the self-query's cross-attention weight per ally slot, tagged by occupant role; slot-ordering bias is ruled out by reversal.

\begin{table}[t]
\centering
\small
\setlength{\tabcolsep}{7pt}
\caption{Cond.~1 role-routing matrix (5 seeds). Entry $(i,j)$ is mean cross-attention weight from role $i$ to ally slots occupied by role $j$, $\pm$ across-seed std. Diagonals are zero by construction.}
\label{tab:routing}
\begin{tabular}{lccc}
\toprule
Agent\,$\backslash$\,Ally & Strike & Vanguard & Support \\
\midrule
Strike   & --- & $0.473{\pm}0.072$ & $\mathbf{0.527{\pm}0.072}$ \\
Vanguard & $\mathbf{0.565{\pm}0.203}$ & --- & $0.435{\pm}0.203$ \\
Support  & $\mathbf{0.633{\pm}0.200}$ & $0.367{\pm}0.200$ & --- \\
\bottomrule
\end{tabular}
\end{table}

The matrix is concentrated, role-specific, and stably off-diagonal. Strike's largest routing weight is on Support, consistent with prediction~4 in $4/5$ seeds at both 3v3 and 9v9. Vanguard and Support both route primarily toward Strike, consistent with the two non-DPS roles attending to the engagement-driver.

\paragraph{MLP gradient attribution.} For Cond.~3 we compute input gradients of the maximum action logit, separately measuring mean magnitude on spatial features and role-onehot dims. The Spatial/Role ratio is $5.8\times$/$2.3\times$/$3.7\times$ (Strike/Vanguard/Support): the MLP registers some role signal but spatial features dominate. The routing-matrix entry std is $0.246$ (Cond.~1) versus $0.055$ for the gradient-based pseudo-routing (Cond.~3), a $4.5\times$ concentration gap. We use entry std as a scale-comparable concentration proxy (zero under uniform routing); row-entropy and KL-from-uniform give the same ranking.

\paragraph{Unified routing metric: occlusion.} Attention weights and MLP gradients are not strictly apples-to-apples, so we additionally measure routing via \emph{occlusion sensitivity} -- zeroing each ally token (mask bit kept at 1) and measuring policy KL from baseline -- which is well-defined for both architectures. On C1 at 3v3, occlusion correlates with attention at row-normalized Pearson $r{=}0.976$ (raw $r{=}0.655$). On C3 at 9v9, occlusion produces erratic magnitudes one to two orders larger than C1 (max-row KL $\approx 5.6$ vs.\ $\approx 0.08$), with the dominant slot varying across seeds rather than being role-conditioned -- the MLP is sensitive to slot contents but not in a role-structured way.

\paragraph{Domain~1 replication (exploratory, 3 seeds).} On MiniGrid, Cond.~1 yields entry std $0.252$, matching SMACv2 ($0.246$). The preferred role pair differs (Vanguard-dominant vs.\ Support-dominant), reflecting different role semantics; the replication target is concentration, not the specific pair.

\paragraph{Cond.~2 routing.} Cond.~2 gives entry std $0.286$, comparable to Cond.~1 and far above Cond.~3: cross-attention alone is sufficient for routing \emph{concentration}. However, Cond.~2 routes Strike$\to$Vanguard in $4/5$ seeds ($0.751$ vs.\ $0.249$), opposite to the role-semantic prior and to Cond.~1, consistent with intra-set blocks influencing which role-pair routing is selected.

\section{Discussion}
\label{sec:discussion}

\paragraph{Summary.} Across two domains, five conditions, and three team sizes, label-conditioned attention yields a stable, role-specific signature ($4.5\times$ more concentrated routing than MLP, positive zero-shot 3v3$\to$9v9 transfer, padding-invariance, $14.0$\,pp from label removal under shared attention) that is robust where win-rate is not: 6v6/9v9 gaps shrink into noise while routing structure persists. The 5-seed re-evaluation aligns 3/4 prior-derived predictions (one tied), reversing two small-$n$ artifacts; Cond.~2 suggests cross-attention supplies \emph{concentration} while intra-set blocks bias \emph{which} role-pair is selected.

\paragraph{Broader implication (speculative).} Beyond the role-pair setting studied here, the diagnostic offers a general template for measuring whether a learned convention is legible from policy architecture -- increasingly relevant as agentic systems coordinate in open-ended environments without designer-specified equilibria, where strategic structure rather than performance alone may determine safety and interpretability. We offer this as a future direction, not a demonstrated result.

\paragraph{Limitations.} Both domains share an entity-centric schema and three roles; SMACv2 evidence is restricted to Terran and 3v3--9v9. Attention weights are not causal explanations; we report them as architecture-exposed patterns corroborated by gradient/occlusion attribution, whose entry-std and occlusion metrics remain heuristic. Role labels are manually injected; deployments without a designer schema must pair the routing pathway with role discovery (e.g., RODE~\cite{rode2021}). The C1 vs.\ C4 contrast bundles label removal with policy sharing, upper-bounding rather than isolating the label effect, and the Cond.~2 ablation also perturbs capacity. Domain~1 Track~2 and 6v6 C1/C3 are 3-seed exploratory runs; all others use 5. We do not claim the learned convention is an equilibrium or optimal under any solution concept, nor that alignment with the prior holds beyond these designer-specified roles -- the contribution is a measurement procedure, not an optimality result.

\section{Conclusion}
\label{sec:conclusion}

We introduced an empirical diagnostic for the coordination conventions selected by cooperative MARL -- a role-routing matrix, formation sensitivity, and a unified occlusion metric -- and used it to compare the learned convention against a theory-informed role-semantic prior. Label-conditioned attention produces a substantially more concentrated, role-specific signature than MLP baselines, stable under scaling and zero-shot 3v3$\to$9v9 transfer. A 5-seed re-evaluation aligns 3/4 prior-derived predictions (one tied) and reveals two earlier divergences as small-$n$ artifacts, underscoring the need for seed-aware structural evaluation. The contribution is a measurement procedure, not an equilibrium or causal claim; future work should extend it to learned role schemas and non-Terran domains.

\acks{The authors would like to thank Professor Jinseok Seo from Dong-Eui University and Gyuyeol Jeong from Netmarble for their insightful discussions and valuable comments that greatly improved this work.}

\bibliography{refs}

\end{document}